\documentclass[12pt]{iopart}

\usepackage{iopams}  

\expandafter\let\csname equation*\endcsname\relax
\expandafter\let\csname endequation*\endcsname\relax

\usepackage{epsfig, amsmath, amssymb}
\newtheorem{theorem}{Theorem}

\newtheorem{assumption}{Assumption}
\usepackage{hyperref}

\usepackage{xcolor}
\usepackage{cite}

\begin{document}

\title[Leveraging Pre-Trained NN to Enhance Machine Learning with VQC]{Leveraging Pre-Trained Neural Networks to Enhance Machine Learning with Variational Quantum Circuits}

\author{Jun Qi$^{1}$, Chao-Han Yang$^{1, 2}$, Samuel Yen-Chi Chen$^{3}$, Pin-Yu Chen$^{4}$, Hector Zenil$^{5}$, Jesper Tegner$^{6}$}

\address{1. School of Electrical and Computer Engineering, Georgia Institute of Technology, Atlanta, GA 30332, USA \\
2. NVIDIA Research, Santa Clara, CA 95051, USA		 \\
3. Wells Fargo, New York, NY 10017, USA  \\
4. IBM Research, Yorktown Heights, NY 10598, USA		 \\
5. Biomedical Engineering and Imaging Sciences, King's College London, London
WC2R 2LS, UK                                             \\
6. Computer, Electrical and Mathematical Sciences and Engineering Division, King
Abdullah University of Science and Technology, Thuwal 23955-6900, Saudi Arabia
}
\ead{jqi41@gatech.edu, jesper.tegner@kaust.edu.sa}
\vspace{10pt}


\begin{abstract}
Quantum Machine Learning (QML) offers tremendous potential but is currently limited by the availability of qubits. We introduce an innovative approach that utilizes pre-trained neural networks to enhance Variational Quantum Circuits (VQC). This technique effectively separates approximation error from qubit count and removes the need for restrictive conditions, making QML more viable for real-world applications. Our method significantly improves parameter optimization for VQC while delivering notable gains in representation and generalization capabilities, as evidenced by rigorous theoretical analysis and extensive empirical testing on quantum dot classification tasks. Moreover, our results extend to applications such as human genome analysis, demonstrating the broad applicability of our approach. By addressing the constraints of current quantum hardware, our work paves the way for a new era of advanced QML applications, unlocking the full potential of quantum computing in fields such as machine learning, materials science, medicine, mimetics, and various interdisciplinary areas.



\end{abstract}

%
%
%
%
%


\section{Introduction}
\label{sec1}

Quantum machine learning (QML) is an emerging interdisciplinary field that integrates the principles of quantum computing and machine learning~\cite{preskill2018quantum, cerezo2022challenges, liu2021rigorous, biamonte2017quantum, schuld2015introduction, schuld2019quantum}. With the rapid development of quantum computing hardware, we are in the noisy intermediate-scale quantum (NISQ) era that admits only a few hundred physical qubits to implement QML algorithms on NISQ devices in the presence of high levels of quantum noise~\cite{egan2021fault, power_data, huggins2019towards}. Variational quantum circuit (VQC) is a promising building block for a QML architecture for processing data and making predictions~\cite{cerezo2021variational, cong2019quantum, benedetti2019parameterized}. The VQC block is composed of parameterized quantum circuits that can be optimized by employing a stochastic gradient descent (SGD) algorithm to minimize a loss function in a black-propagation manner~\cite{beer2020training, sharma2022trainability, du2020expressive, huembeli2021characterizing}. The VQC's resilience against quantum noise errors admits it to be applicable in many QML applications~\cite{yang2021decentralizing, huang2021experimental, chen2022quantumCNN, chen2020variational, caro2022generalization, liu2024towards, qi2024federated, mcclean2016theory, li2017hybrid, callison2022hybrid}. 

However, the number of available qubits constrains the VQC's representation power, and the amount of training data limits its generalization power~\cite{qi2020analyzing, du2021learnability, mitarai2018quantum}. Based on pre-trained neural networks, our theoretical findings introduce new bounds that allow the representation power independent of the number of qubits. We enhance the generalization power by requiring only a smaller target dataset. Notably, this target dataset pertains to quantum data that is significantly different from the source data used to train a classical neural network. Thus, this work focuses on both theoretical and empirical advancements in using pre-trained neural networks for VQC, including:

\begin{enumerate}
\item Demonstrating that pre-trained neural networks can enhance the representation and generalization powers of VQC blocks.
\item Validating our theoretical insights with experimental results from semiconductor quantum dot classification and human genome transcription factor binding site (TFBS) prediction.
\end{enumerate}

Unlike the end-to-end learning approach in hybrid quantum-classical architectures, we keep the pre-trained neural network's parameters fixed without further tuning. This shows that this can enhance the VQC's representation and generalization capabilities. This strategy allows us to utilize large pre-trained neural networks, including cutting-edge large language models with frozen parameters, to scale up QML using VQC blocks. More importantly, we are the first to establish the theoretical benefits of using pre-trained neural networks for VQC and to pioneer their application in quantum data scenarios.

\begin{figure}
\centerline{\epsfig{figure=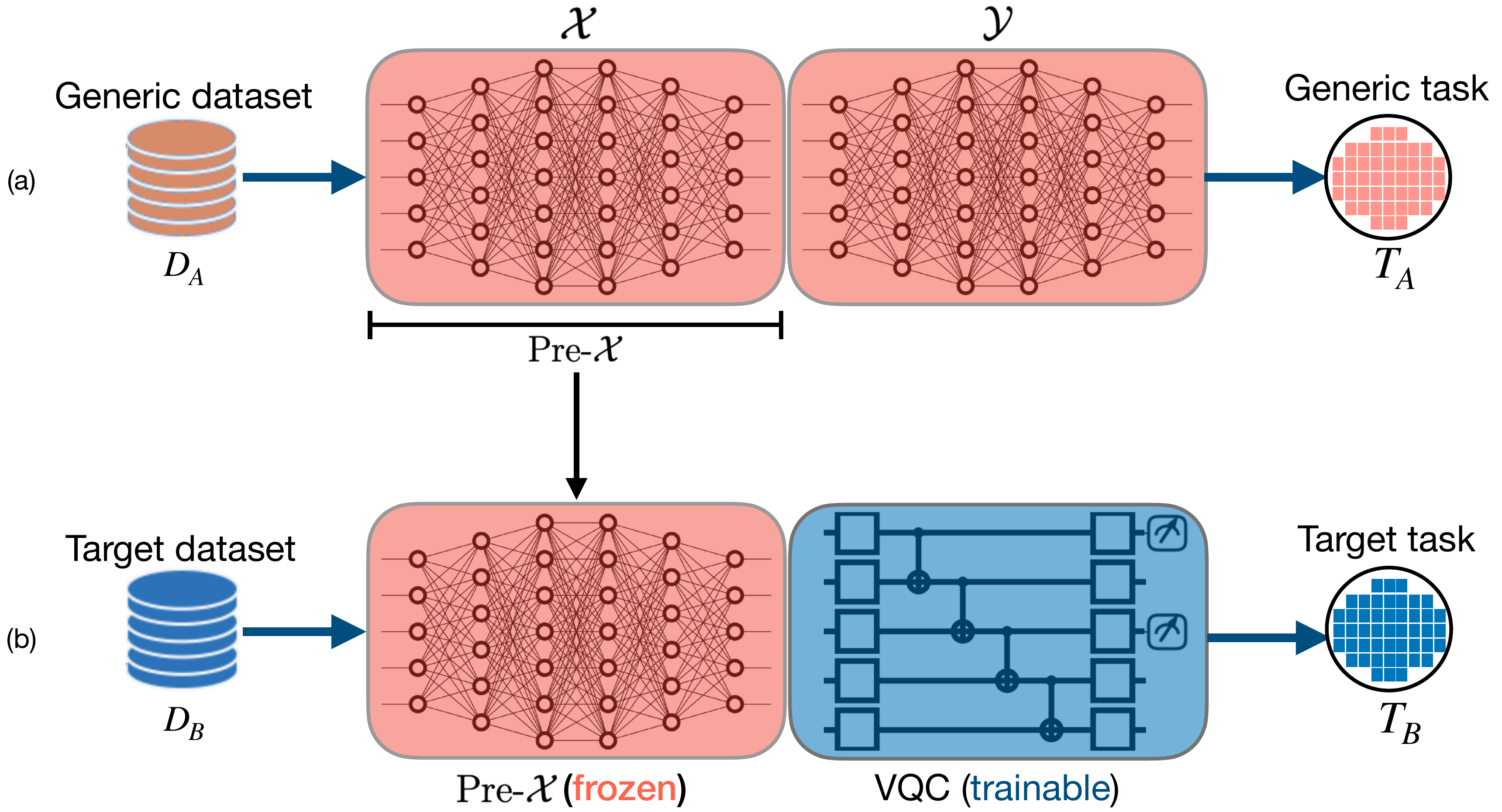, width=110mm}}
\caption{{\textbf{The architecture of using pre-trained neural networks for VQC}. (a) A classical model $\mathcal{X}$ is pre-trained with another classical one $\mathcal{Y}$ on a generic dataset $D_{A}$ for a generic task $T_{A}$. (b) the pre-trained classical model Pre-$\mathcal{X}$ is transferred to a hybrid quantum-classical architecture, where the classical model Pre-$\mathcal{X}$ is frozen without a further parameter adjustment, and the VQC's parameters need to be trained based on a target dataset $D_{B}$ for a target task $T_{B}$.}}
\label{fig:transfer}
\end{figure}

We show the architecture of pre-trained neural networks for VQC in Figure~\ref{fig:transfer}. A classical neural network $\mathcal{X}$ is pre-trained with another neural network $\mathcal{Y}$ following an end-to-end learning pipeline on a generic dataset $D_{A}$ for a generic task $T_{A}$. The well-trained neural network Pre-$\mathcal{X}$ is then transferred to a hybrid quantum-classical model comprised of classical module Pre-$\mathcal{X}$ and a quantum component VQC. In the fine-tuning stage, only the VQC's parameters need further update on a target dataset $D_{B}$ given a target task $T_{B}$. Besides, the pre-trained neural network is closely related to the state-of-the-art foundation model of generative artificial intelligence, e.g., large language models like Generative Pre-trained Transformers (GPT)~\cite{floridi2020gpt, DevlinCLT19}, which can generate context-enriched features before going through the VQC block. 

Recent studies have shown that hybrid quantum-classical models have achieved remarkable empirical results in numerous machine-learning applications. These models were developed using an end-to-end learning approach, where the parameters of both classical and quantum models are jointly optimized with the training data. However, in our work, we keep the parameters of the classical neural network fixed and only allow the VQC's parameters to be trainable. Specifically, we perform an error performance analysis on our proposed method, demonstrating that pre-trained neural networks can enhance the VQC's representation and generalization capabilities.

Based on the error performance analysis, a generalized loss can be decomposed into the sum of three error components: approximation error, estimation error, and optimization error. The approximation error is associated with the representation power, and the estimation and optimization errors jointly correspond to the generalization power. By separately providing upper bounds on the three error terms, we offer a theoretical foundation on the algorithm of a pre-trained neural network for VQC, which exhibits theoretical improvement to the VQC model by comparing our newly derived upper bounds with the established ones for VQC. Our theoretical results are shown in Table~\ref{tab:comp} and summarized as follows: 

\begin{table}[t]\footnotesize
\center
\renewcommand{\arraystretch}{1.3}
\caption{Summarizing our theoretical results of using pre-trained neural networks for VQC (abbreviated as Pre-$\mathcal{X}$+VQC) and comparing them with VQC in~\cite{qi2023theoretical}.}
\begin{tabular}{|c||c|c|}
\hline
	\textbf{Error Component}	&	Pre-$\mathcal{X}$+VQC	&	VQC		 \\
\hline
Approximation Error		&   $\tilde{\mathcal{O}}\left(\sqrt{\frac{C(\mathbb{F}_{\mathcal{X}})}{\vert D_{A} \vert}} \right) + \mathcal{O}\left(\frac{1}{\sqrt{M}}\right)$	&  $\tilde{\mathcal{O}}\left(\frac{1}{\sqrt{U}}\right) + \mathcal{O}\left(\frac{1}{\sqrt{M}}\right)$	\\
\hline
Estimation Error	 	&  $\tilde{\mathcal{O}}\left(\sqrt{\frac{C(\mathbb{F}_{V})}{\vert D_{B} \vert}}\right)$	&  $ \tilde{\mathcal{O}}\left(\sqrt{\frac{C(\mathbb{F}_{V})}{\vert D \vert}}\right)  $ 	\\
\hline
Conditions on Optimization Error	&  Pre-trained NNs for VQC		&  PL assumption	\\
\hline
Optimization Error		& $\beta R^{2} + R\sqrt{\frac{L^{2} + \beta^{2} R^{2}}{T_{\rm sgd}}}$  	&	$ \approx 0$		\\
\hline
\end{tabular}
\label{tab:comp}
\end{table}

\begin{itemize}
\item Approximation error: we provide an error upper bound on the approximation error with the form as $\tilde{\mathcal{O}}\left(\sqrt{\frac{C(\mathbb{F}_{\mathcal{X}})}{\vert D_{A} \vert}} \right) + \mathcal{O}\left(\frac{1}{\sqrt{M}}\right)$, where $\mathbb{F}_{\mathcal{X}}$ denotes the functional class of pre-trained neural network Pre-$\mathcal{X}$, $C(\cdot)$ is the measurement of the intrinsic complexity of the functional class, $\vert D_{A} \vert$ and $M$ refer to the amount of training data and the counts of quantum measurement. Compared with our previous upper bound on the VQC's approximation error, our newly derived upper bound is independent of the number of qubits $U$, allowing for a few qubits in practice to achieve a small approximation error by scaling down the term $\sqrt{\frac{C(\mathbb{F}_{\mathcal{X}})}{\vert D_{A} \vert}}$. 
\item Estimation error: our upper bound on the estimation error is taken in the form as $\tilde{\mathcal{O}}\left(\sqrt{\frac{C(\mathbb{F}_{V})}{\vert D_{B} \vert}}\right)$, where $C(\cdot)$ also denotes the measurement for the complexity of functional class, $\mathbb{F}_{V}$ refers to the functional class of VQC and $\vert D_{B} \vert$ denotes the number of target data for the VQC training. Our newly derived upper bound is reduced to a small value with increasing target data in $D_{B}$. In comparison with our previous upper bound on the VQC's estimation error, the target data can be limited to a small scale to obtain a well-trained VQC model, which means that the number of target data in $D_{B}$ could be much smaller than the amount of training data in $D$ for VQC. 
\item Optimization error: we demonstrate that the SGD algorithm returns an upper bound as $\beta R^{2} + R\sqrt{\frac{L^{2} + \beta^{2} R^{2}}{T_{\rm sgd}}}$, where $R$, $\beta$, and $L$ are pre-defined hyper-parameters and $T_{\rm sgd}$ denotes the number of epochs in the VQC training process. Unlike the necessary setup of Polyak-{\L}ojasiewicz (PL) condition~\cite{karimi2016linear} for VQC to ensure a small optimization error, our proposed QML approach shows that it does not require such a prior condition. By controlling the hyperparameters $R$, $\beta$, and $L$, we can establish a connection between the optimization error and the constraints for the SGD algorithm. 
\end{itemize}

Besides, we assess and confirm the effectiveness of pre-trained neural networks for VQC through a practical use case involving the classification of semiconductor quantum dots and the prediction of transcription factor binding sites (TFBS) in the human genome. QDs are promising candidates for creating qubits, which are the fundamental components of NISQ devices. Our experiments focus on QD autotuning, which involves identifying charge state transition lines in two-dimensional stability diagrams for binary class pattern classification~\cite{de2002intrinsic, brennan2011atomic, tang2015storage}. For TFBS prediction, the task relates to proteins that regulate gene expression by binding to specific DNA regions, such as promoters, and either activating or repressing gene expression. Each transcription factor has a distinct binding motif, a TFBS pattern. This prediction task involves quantum data derived from DNA sequences, making it well-suited for our QML paradigm~\cite{tompa2005assessing, ou2018motifstack, berger2006compact}.


\section{Results}

\subsection{Preliminaries}

\begin{figure}
\centerline{\epsfig{figure=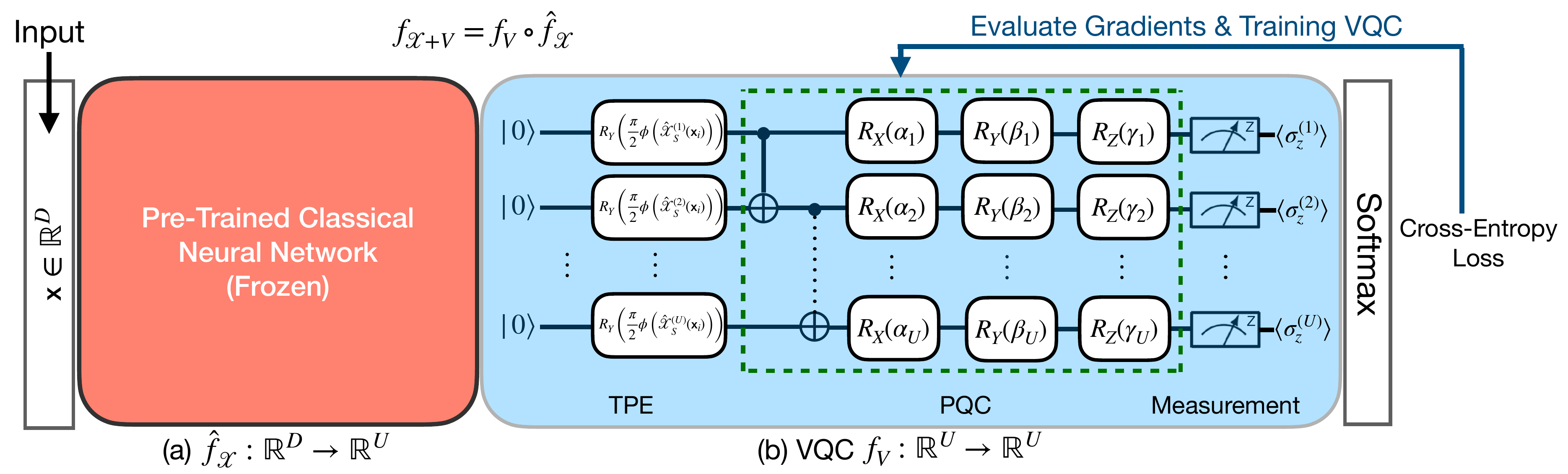, width=150mm}}
\caption{\textbf{Illustration of hybrid quantum-classical architecture with a pre-trained classical neural network and a VQC}. Given an input vector $\textbf{x}\in \mathbb{R}^{D}$, a pre-trained classical neural network transforms the input into a context-enriched feature representation $\hat{f}_{\mathcal{X}}(\textbf{x})$, which is transformed into classical outputs $\langle \sigma_{z}^{(i)} \rangle$ through the VQC block. A softmax operation is utilized for pattern classification, where a cross-entropy loss is used to calculate the gradients to update the VQC's parameters.}
\label{fig:hybrid}
\end{figure}

Before delving into our theoretical and empirical results, we first introduce the VQC architecture representing the QML model. As shown in component (b) of Figure~\ref{fig:hybrid}, the VQC block consists of three components: (a) Tensor Product Encoding (TPE), (b) Parametric Quantum Circuit (PQC), and (c) Measurement. 

Given $U$ qubits, we employ $U$ Pauli-Y gates $R_{Y}(\cdot)$ to constitute a TPE operation, which transforms a classical input vector $\textbf{x} = [x_{1}, x_{2}, ..., x_{U}]^{\top}$ into their corresponding quantum state $\vert \textbf{x} \rangle = [\vert x_{1} \rangle, \vert x_{2} \rangle, ..., \vert x_{U} \rangle]^{\top}$ through adopting a one-to-one mapping as:
\begin{equation}
\label{eq:tpe}
\vert \textbf{x} \rangle = \left(\bigotimes_{i=1}^{U} R_{Y}(\frac{\pi}{2} \phi(x_{i})) \right) \vert 0 \rangle^{\otimes U} = \begin{bmatrix} \cos(\frac{\pi}{2} \phi(x_{1})) \\ \sin(\frac{\pi}{2} \phi(x_{1})) \end{bmatrix} \otimes \begin{bmatrix} \cos(\frac{\pi}{2} \phi(x_{2})) \\ \sin(\frac{\pi}{2} \phi(x_{2})) \end{bmatrix} \otimes \cdot\cdot\cdot \otimes \begin{bmatrix} \cos(\frac{\pi}{2} \phi(x_{U})) \\ \sin(\frac{\pi}{2} \phi(x_{U})) \end{bmatrix},
\end{equation}
where $\phi(\cdot)$ is a non-linear function like a sigmoid function $\phi(x_{i}) = \frac{1}{1 + \exp(-x_{i})}$ such that $\phi(x_{i})$ is restricted to the domain of $[0, 1]$ and we can establish a one-to-one mapping between $\textbf{x}$ and $\vert \textbf{x} \rangle$. 

In the PQC framework, we first implement quantum entanglement through a series of two-qubit controlled-not (CNOT) gates and then leverage single-qubit Pauli rotation gates $R_{X}(\alpha_{i})$, $R_{Y}(\beta_{i})$ and $R_{Z}(\gamma_{i})$ with trainable parameters of qubit rotation angles $\alpha_{i}$, $\beta_{i}$, $\gamma_{i}$. The PQC model in the green dashed square is repeatedly copied to build up a deep PQC architecture, which outputs $U$ quantum states $\vert o_{1} \rangle$, $\vert o_{2} \rangle$, ..., $\vert o_{U} \rangle$. The quantum measurement returns the expected values $\langle \sigma_{z}^{(i)} \rangle = \langle o_{i} \vert \sigma_{z}^{(i)} \vert o_{i} \rangle$ associated with the Pauli-Z operators $\sigma_{z}^{(i)}$ for the $i^{\rm th}$ quantum channel. The expected values $\langle \sigma_{z}^{(i)} \rangle = \langle o_{i} \vert \sigma_{z}^{(i)} \vert o_{i} \rangle$ can be approximated by using an arithmetic average of $M$ times' quantum measurement. The resulting values $\langle \sigma_{z}^{(i)} \rangle$ are connected to a softmax operation with a cross-entropy loss to calculate gradients for the update of VQC's parameters.

\subsection{Theoretical results}

Figure~\ref{fig:hybrid} illustrates Pre-$\mathcal{X}$+VQC with a hybrid quantum-classical architecture that consists of a pre-trained classical neural network and a VQC block, which are separately shown in components (a) and (b). The pre-trained classical neural network transforms $D$-dimensional input vectors $\textbf{x} = [x_{1}, x_{2}, ..., x_{D}]^{\top}$ into context-enriched features $\hat{f}_{\mathcal{X}}(\textbf{x})$. The features $\hat{f}_{\mathcal{X}}(\textbf{x})$ are then turned into quantum states $R_{Y}(\frac{\pi}{2} \phi(\hat{f}_{\mathcal{X}}(x_{i})))\vert 0 \rangle$ through the TPE operation in the VQC block. The VQC block transforms the quantum states into their corresponding classical outputs $\langle \sigma_{z}^{(i)} \rangle$ that are connected to a softmax operation with the cross-entropy loss. The VQC's parameters are adjustable using the SGD algorithm to adapt to the input training data. In contrast, the pre-trained neural network's parameters are frozen without participating in the VQC fine-tuning process. 

Next, we characterize the Pre-$\mathcal{X}$ +VQC's representation and generalization powers. Mathematically, given a functional class of neural networks $\mathbb{F}_{\mathcal{X}}$ and a VQC functional class $\mathbb{F}_{V}$, we define a hybrid quantum-classical model as $f_{\mathcal{X} + V} = f_{V} \circ \hat{f}_{\mathcal{X}}$, where a pre-trained neural network $\hat{f}_{\mathcal{X}} \in \mathbb{F}_{\mathcal{X}}$ and $f_{V} \in \mathbb{F}_{V}$. Additionally, we separately use the symbols $\bar{f}$, $\hat{f}$, and $f^{*}$ to represent the empirical risk minimizer, actual algorithm-returned hypothesis, and optimal hypothesis. Given an underlying distribution $\mathcal{D}$, for a smooth target operator $h_{\mathcal{D}}^{*}$ and a loss function $\ell$ to measure the distance between $h_{\mathcal{D}}^{*}$ and an operator $f$, we define an expected loss as:
\begin{equation}
\mathcal{L}_{\mathcal{D}}(f) := \textbf{E}_{\textbf{x} \sim \mathcal{D}}\left[	\ell(h^{*}_{\mathcal{D}}(\textbf{x})), f(\textbf{x}) \right]. 
\end{equation}

We further assume that the training dataset $S$ contains the samples $\{\textbf{x}_{1}, \textbf{x}_{2}, ..., \textbf{x}_{\vert S\vert}\}$, the expected loss $\mathcal{L}_{\mathcal{D}}(f)$ can be approximated by an empirical loss $\mathcal{L}_{S}(f)$ as: 
\begin{equation}
\mathcal{L}_{S}(f) := \frac{1}{\vert S \vert} \sum\limits_{n=1}^{\vert S \vert} \ell(h_{\mathcal{D}}^{*}(\textbf{x}_{n}), f(\textbf{x}_{n})). 
\end{equation}

Then, we conduct an error performance analysis by decomposing the expected loss into the sum of approximation error, estimation error, and optimization error, which is 
\begin{equation}
\begin{split}
\label{eq:f}
\mathcal{L}_{\mathcal{D}}(\hat{f}_{\mathcal{X} + V}) &=  \underbrace{\mathcal{L}_{\mathcal{D}}(f_{\mathcal{X} + V}^{*})}_{\textcolor{blue}{\text{Approximation\hspace{0.5mm} Error}}}  +  \underbrace{\mathcal{L}_{\mathcal{D}}(\bar{f}_{\mathcal{X} + V}) - \mathcal{L}_{\mathcal{D}}(f_{\mathcal{X} + V}^{*})}_{\textcolor{blue}{\text{Estimation\hspace{0.5mm} Error}}} + \underbrace{ \mathcal{L}_{\mathcal{D}}(\hat{f}_{\mathcal{X} + V}) - \mathcal{L}_{\mathcal{D}}(\bar{f}_{\mathcal{X} + V})}_{\textcolor{blue}{\text{Optimization\hspace{0.5mm} Error}}} \\
&= \epsilon_{\rm app} + \epsilon_{\rm est} + \epsilon_{\rm opt},
\end{split}
\end{equation}
where the Pre-$\mathcal{X}$ +VQC's optimal hypothesis $f_{\mathcal{X} + V}^{*} = \arg\min_{f_{V}\in \mathbb{F}_{V}} \mathcal{L}_{\mathcal{D}}(f_{V} \circ \hat{f}_{\mathcal{X}})$ and its empirical risk minimizer $\bar{f}_{\mathcal{X} + V} = \arg\min_{f_{V}\in \mathbb{F}_{V}} \mathcal{L}_{S}(f_{V} \circ \hat{f}_{\mathcal{X}})$, and the symbols $\epsilon_{\rm app}$, $\epsilon_{\rm est}$ and $\epsilon_{\rm opt}$ denote approximation error, estimation error, and optimization error, respectively. 

We first upper bound the approximation error associated with its representation power as presented in Theorem~\ref{thm:thm1}, and then we derive the upper bounds on the estimation error and the optimization error as shown in Theorem~\ref{thm:thm2} and \ref{thm:thm3} that jointly correspond to the generalization power. 

\begin{theorem}
\label{thm:thm1}
Given a smooth target operator $h_{\mathcal{D}}^{*}$ and $U$ quantum channels for the VQC model, for a pre-trained neural network $\hat{f}_{\mathcal{X}} \in \mathbb{F}_{\mathcal{X}}$ conducted on a source training dataset $D_{A}$, we can find a VQC model $f_{V} \in \mathbb{F}_{V}$ such that for a cross-entropy loss function $\mathcal{L}_{\mathcal{D}}: \mathbb{R}^{U} \rightarrow \mathbb{R}$, there exists a Pre-$\mathcal{X}$+VQC operator $f^{*}_{\mathcal{X}+V} = f^{*}_{V} \circ \hat{f}_{\mathcal{X}}$ satisfying 
\begin{equation}
\label{eq:app}
\epsilon_{\rm app} = \mathcal{L}_{\mathcal{D}}(f^{*}_{\mathcal{X}+V}) = \tilde{\mathcal{O}}\left(\sqrt{\frac{C(\mathbb{F}_{\mathcal{X}})}{\vert D_{A} \vert}} \right) + \mathcal{O}\left(\frac{1}{\sqrt{M}}\right), 
\end{equation}
where $M$ is the counts of quantum measurement, $C(\cdot)$ is the measurement of the intrinsic complexity of the functional class, and $\mathbb{F}_{\mathcal{X}}$ is associated with the functional class of classical neural networks $\mathcal{X}$. 
\end{theorem}

Theorem~\ref{thm:thm1} suggests that the upper bound on the approximation error consists of two parts: the transfer learning risk from the pre-trained neural network $\mathcal{X}$ and the counts of quantum measurement. An enormous source dataset $D_{A}$ and more counts of the quantum measurement $M$ are expected to lower the approximation error of Pre-$\mathcal{X}$+VQC. In particular, compared with our previous theoretical result of VQC, the number of qubits $U$ does not show up in the upper bound for the approximation error. Since only the number of quantum measurements matters for the Pre-$\mathcal{X}$ +VQC's representation capability, our QML approach can alleviate the necessity to pursue many qubits to enhance the VQC's representation power. 

\begin{theorem}
\label{thm:thm2}
Based on the setup of Pre-$\mathcal{X}$+VQC in Theorem~\ref{thm:thm1}, given the target dataset $D_{B}$, the estimation error is upper bounded as:
\begin{equation}
\epsilon_{\rm est} = \mathcal{L}_{\mathcal{D}}(\bar{f}_{\mathcal{X}+V}) - \mathcal{L}_{\mathcal{D}}(f_{\mathcal{X}+V}^{*}) = \tilde{\mathcal{O}}\left(\sqrt{\frac{C(\mathbb{F}_{V})}{\vert D_{B} \vert}}\right), 
\end{equation}
where $\mathbb{F}_{V}$ denotes the VQC's functional class, $C(\cdot)$ represents the measurement of the intrinsic complexity of the functional class, and $D_{B}$ is the target dataset. 
\end{theorem}

The upper bound on the estimation error $\epsilon_{\rm est}$ relies on the term $C(\mathbb{F}_{V})$ that measures the intrinsic complexity of the VQC's functional class $\mathbb{F}_{V}$ and the size of the dataset $D_{B}$. The dataset $D_{B}$ can be as large as $C(\mathbb{F}_{V})$ to obtain a small $\epsilon_{\rm est}$. Theorem~\ref{thm:thm2} suggests that the estimation error can be reduced to an arbitrarily small value if the amount of target dataset $D_{B}$ is sufficiently large. 

To derive our upper bound on the optimization error, we first make two Assumptions: Approximate Linearity and Gradient Bound, which are presented below: 

\begin{assumption}[Approximate Linearity]
\label{ass1}
Let $D_{B} = \{\textbf{x}_{1}, \textbf{x}_{2}, ..., \textbf{x}_{\vert D_{B} \vert}\}$ be the target dataset. There exists a $f_{\mathcal{X}+V} = f_{V} \circ \hat{f}_{\mathcal{X}}$ with a VQC's parameter vector $\boldsymbol{\theta}$, denoted as $f_{\mathcal{X}+V}(\textbf{x}_{n}; \boldsymbol{\theta})$. Then, for two constants $\beta$, $L$ and a first-order gradient $\boldsymbol{\delta}$, we have 
\begin{equation}
\sup\limits_{\boldsymbol{\delta}} \frac{1}{\vert D_{B}\vert} \sum\limits_{n=1}^{\vert D_{B} \vert} \left\lVert  \nabla_{\boldsymbol{\theta}}^{2} f_{\mathcal{X}+V}(\textbf{x}_{n}; \boldsymbol{\theta} + \boldsymbol{\delta}) \right\rVert_{2}^{2} \le \beta^{2}, 
\end{equation}
and 
\begin{equation}
\sup\limits_{\boldsymbol{\delta}} \frac{1}{\vert D_{B} \vert} \sum\limits_{n=1}^{\vert D_{B} \vert} \left\lVert \nabla_{\boldsymbol{\theta}}f_{\mathcal{X} + V}(\textbf{x}_{n}; \boldsymbol{\theta}) \right\rVert_{2}^{2} \le L^{2}. 
\end{equation}
\end{assumption}

\begin{assumption}[Gradient Bound]
\label{ass2}
For the first-order gradient $\boldsymbol{\delta}$ of the VQC's parameters in the training process, we assume that $\left\lVert \boldsymbol{\delta} \right\rVert_{2}^{2} \le R$ for a constant $R$. 
\end{assumption}

Then, we can derive the upper bound on the optimization error $\epsilon_{\rm opt}$ as shown in Theorem~\ref{thm:thm3}, which suggests that the upper bound on $\epsilon_{\rm opt}$ is controlled by the constant factor $R$ for the supremum value $\lVert \boldsymbol{\delta} \rVert_{2}^{2}$. 

\begin{theorem}
\label{thm:thm3}
Given the constants $\beta$, $R$, and $L$ in Assumptions~\ref{ass1} and~\ref{ass2}, for a total of $T_{\rm sgd}$ iterations with a learning rate $\eta$ as:
\begin{equation}
\eta = \frac{1}{\sqrt{T_{\rm sgd}}} \left( \frac{R}{\sqrt{L^{2} + \beta^{2} R^{2}}} \right), 
\end{equation}
We have: 
\begin{equation}
\epsilon_{\rm opt} \le \beta R^{2} + R \sqrt{\frac{L^{2} + \beta^{2} R^{2}}{T_{\rm sgd}}}. 
\end{equation}
\end{theorem}

Compared with the VQC model with PL condition to guarantee a small optimization error, our QML approach ensures an upper bound related to the iteration step $T_{\rm sgd}$ and the constraints constant $R$ for the supremum value of $\Vert \boldsymbol{\delta} \rVert_{2}^{2}$. To reduce the upper bound of $\epsilon_{\rm opt}$, a significant value $T_{\rm sgd}$ and a small constant $R$ are expected to obtain a small learning rate $\eta$.

\subsection{Empirical results of quantum dots classification}

\begin{figure}
\centerline{\epsfig{figure=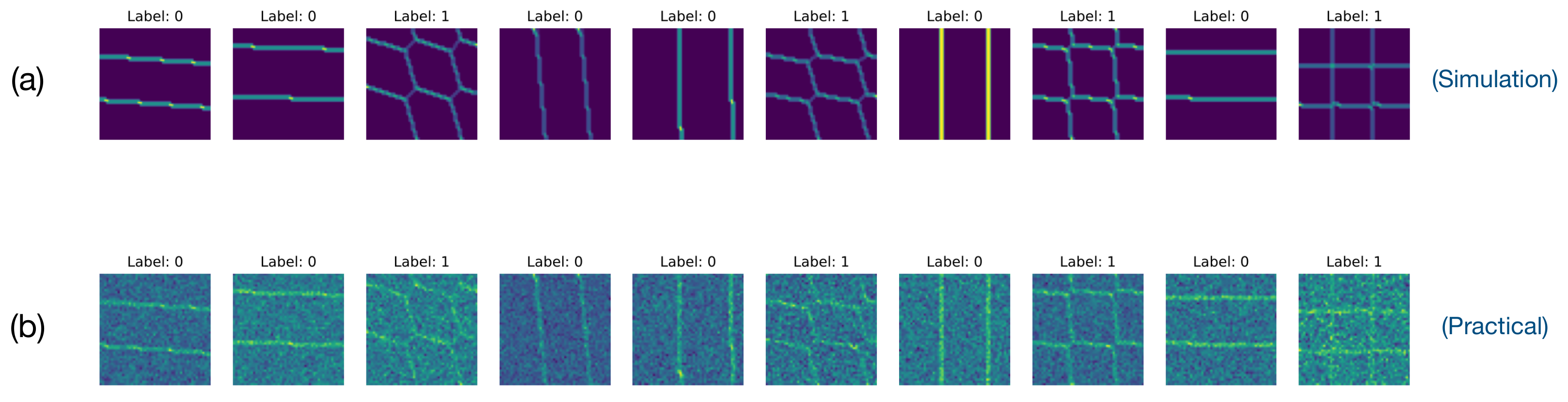, width=140mm}}
\caption{\textbf{Illustration of single and double quantum dot charge stability diagrams}. (a) labeled clean charge stability diagrams containing the transition lines without noise; (b) labeled noisy charge stability diagrams mixed with realistic noise effects on the transition lines. Label:$0$ and Label:$1$ denote charge stability diagrams of single and double quantum dots. By detecting the transition lines using the QML approach, we aim to judge whether the charge stability diagram corresponds to the single or double quantum dots. In particular, we use the noiseless data to evaluate the models' representation power while using the noisy data to assess the models' generalization power.}
\label{fig:dot}
\end{figure}

To corroborate our theoretical results, we conduct our experimental simulation on the binary classification of single and double quantum dot charge stability diagram. As shown in Figure~\ref{fig:dot}, clean and noisy charge stability diagrams separately contain the transition lines with and without realistic noisy effects. The charge stability diagrams with Label $0$ and $1$ correspond to single and double quantum dots, respectively. This work applies the QML approach to classify the charge stability diagram's class. 

Our experimental simulations are divided into two groups: (1) assessing the representation power of Pre-$\mathcal{X}$+VQC, both training and test data are set in the same clean environment as shown in Figure~\ref{fig:dot} (a); (2) evaluating the generalization power of Pre-$\mathcal{X}$+VQC, we leverage the noisy data with realistic noisy effects, as shown in Figure~\ref{fig:dot} (b). The training data associated with the charge stability diagrams are employed to update the VQC's parameters, and the pre-trained neural network Pre-$\mathcal{X}$ is frozen without further fine-tuning procedure. We compose the Pre-$\mathcal{X}$ model with either ResNet18 or ResNet50, where both ResNet18 and ResNet50 are classical CNN modules pre-trained on an extensive database of ImageNet~\cite{deng2009imagenet}. Accordingly, the resulting hybrid quantum-classical model is separately denoted as Pre-ResNet18+VQC and Pre-ResNet50+VQC. Our baseline model, by contrast, is built upon a VQC pipeline. Moreover, to highlight the performance advantages of our QML approach, we compare our proposed methods with their classical counterparts, Pre-ResNet18+NN and Pre-ResNet50+NN, respectively, where NN represents a classical feed-forward neural network with one hidden layer of dimensions being equivalent to the number of qubits. We design the experimental quantum dots classification to verify the following points: 

\begin{itemize}
\item The Pre-$\mathcal{X}$+VQC model can achieve better representation and generalization powers than VQC, consistent with our theoretical results. 
\item The number of qubits is unrelated to the representation power of Pre-$\mathcal{X}$+VQC, implying that the increase of qubits cannot boost its representation capability. 
\item The pre-trained neural networks can help VQC to achieve exponential convergence rates in the VQC training process. 
\end{itemize}

As demonstrated in Figure~\ref{fig:dot}, the dataset is composed of a simulated $50 \times 50$ squared image lattice for both noisy and noiseless charge stability diagrams of quantum dots, which contain $2,000$ noiseless data and $2,000$ noisy ones. We randomly partition the noiseless data into $1,800$ training data and $200$ test ones to evaluate the models' representation power. To assess the models' generalization power, we randomly assign $1,800$ noisy data into the training set and put the remaining $200$ into the test set. 

In our experimental setup, we employ $8$ qubits, corresponding to $8$ quantum channels, to compose a VQC model and set the PQC's depth and learning rate as $2$ and $0.001$, respectively. We select the first two measurement outcomes $\langle \sigma_{z}^{(1)} \rangle$ and $\langle \sigma_{z}^{(2)} \rangle$ as the outputs connected to the softmax operation for the binary classification. The cross-entropy is taken as the loss function to calculate the loss values and return first-order gradients to update the VQC's parameters. At the same time, we freeze the ResNet's parameters without further fine-tuning. To build our VQC baseline system, a principle component analysis (PCA)~\cite{abdi2010principal} is utilized to reduce the dimensionality of classical inputs to $8$, matching the number of the VQC's quantum channels. 

\begin{figure}
\centerline{\epsfig{figure=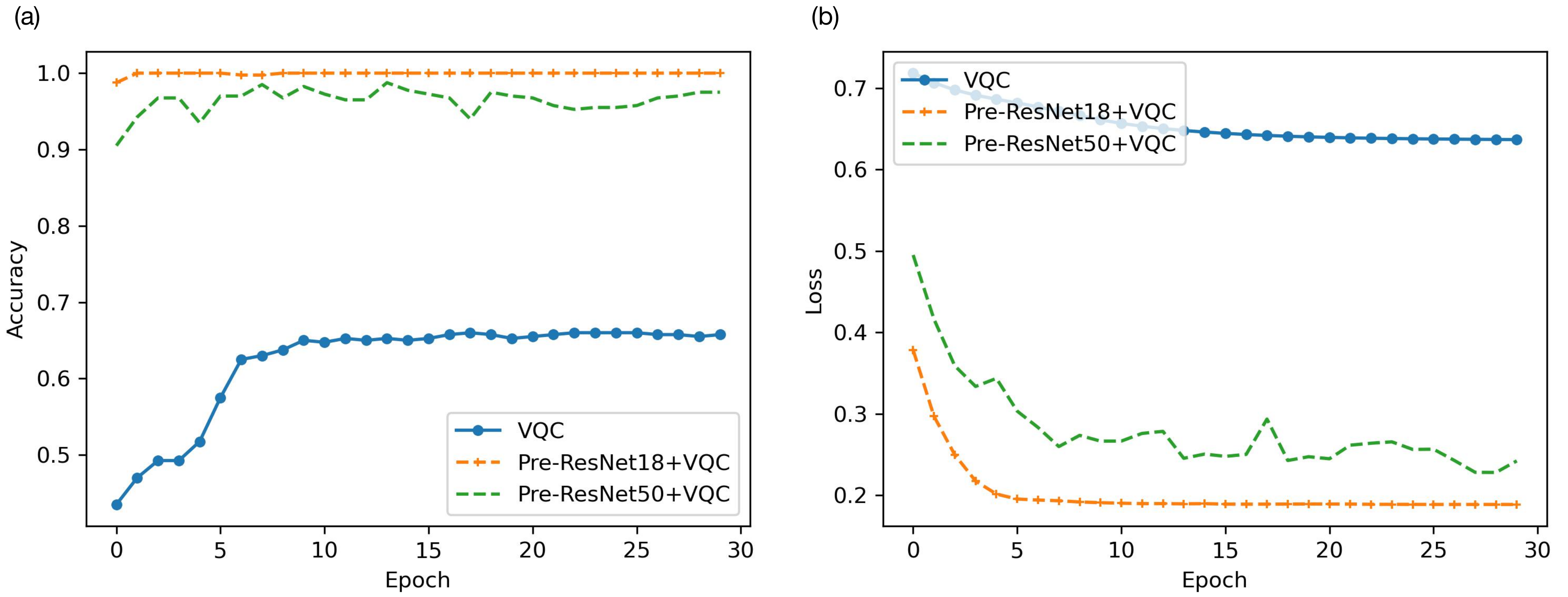, width=140mm}}
\caption{\textbf{Experimental results on noiseless charge stability diagrams for evaluating the models' representation powers with $8$ qubits}. (a) The experimental results are based on accuracy; (b) The measurement of the models' performance is conducted through the loss values. Both Pre-ResNet18+VQC and Pre-ResNet50+VQC can achieve better empirical results than VQC in terms of accuracy and loss values, and Pre-ResNet18+VQC even outperforms Pre-ResNet50+VQC on the clean dataset.}
\label{fig:dot_rep}
\end{figure}

Figure~\ref{fig:dot_rep} demonstrates our experimental results to assess the representation powers of VQC, Pre-ResNet18+VQC, and Pre-ResNet50+VQC. Our experiments show that both Pre-ResNet18+VQC and Pre-ResNet50+VQC exhibit much better empirical results than the VQC counterpart in terms of higher classification accuracy and lower convergence loss values, which exactly corroborate our theoretical results that the pre-trained ResNets can improve the VQC's representation power using the hybrid quantum-classical architecture. Furthermore, Pre-ResNet18+VQC can consistently outperform Pre-ResNet50+VQC over all of the epochs, achieving a higher accuracy ($99.85\%$ vs. $97.75\%$) and a lower cross-entropy loss scores (0.1886 vs. 0.2422) at the end of a final epoch. The empirical results precisely match our theoretical result for the representation power, as shown in Eq. (\ref{eq:app}), that a more complex pre-trained neural network corresponds to a more prominent upper bound on the approximation error of Pre-$\mathcal{X}$+VQC. Since the model complexity for ResNet$18$ is smaller than ResNet$50$, it is reasonable for Pre-ResNet18+VQC to exhibit a better representation power than Pre-ResNet50+VQC. 

\begin{figure}
\centerline{\epsfig{figure=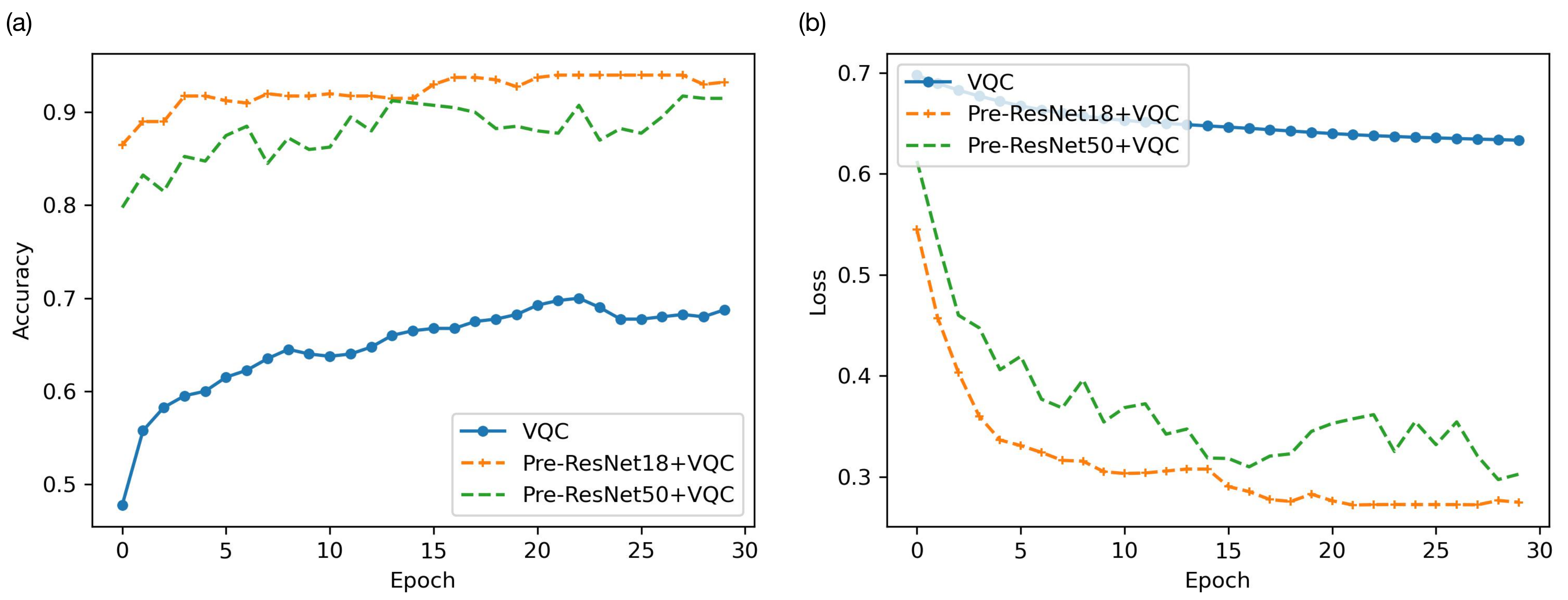, width=140mm}}
\caption{\textbf{Experimental results on noisy charge stability diagrams for assessing VQC models' generalization powers with $8$ qubits}. (a) The experimental results are based on accuracy; (b) The measurement of the models' performance is conducted through the loss values. Both Pre-ResNet18+VQC and Pre-ResNet50+VQC obtain better empirical performance than VQC in terms of accuracy and loss values, and they eventually achieve a very close result on the noisy dataset.}
\label{fig:dot_gen}
\end{figure}

Moreover, Figure~\ref{fig:dot_gen} shows that both Pre-ResNet18+VQC and Pre-ResNet50+VQC obtain significantly lower cross-entropy losses and higher accuracies than VQC, corroborating our theoretical result that a pre-trained neural network can enhance the VQC's generalization power. Although Pre-ResNet18+VQC outperforms Pre-ResNet50+VQC on the noisy dataset, our experimental results suggest that its relative marginal gain is smaller than the results in Figure~\ref{fig:dot_rep}. The empirical results correspond to our theoretical results on the generalization power, where we demonstrate that the upper bound on the estimation error relies upon the model complexity of VQC and the amount of target data, and the generalization performance difference comes from the optimization error. Most importantly, the pre-trained neural networks, ResNet$18$ and ResNet$50$, alleviate the setup of the PL condition for VQC to attain an outstanding optimization performance. Besides, Figure~\ref{fig:dot_rep} (b) and Figure~\ref{fig:dot_gen} (b) suggest that both Pre-ResNet18+VQC and Pre-ResNet50+VQC exhibit exponential convergence rates of cross-entropy losses. It corresponds to the point that the pre-trained neural network can enhance the optimization performance in the models' training stage. 

\begin{figure}
\centerline{\epsfig{figure=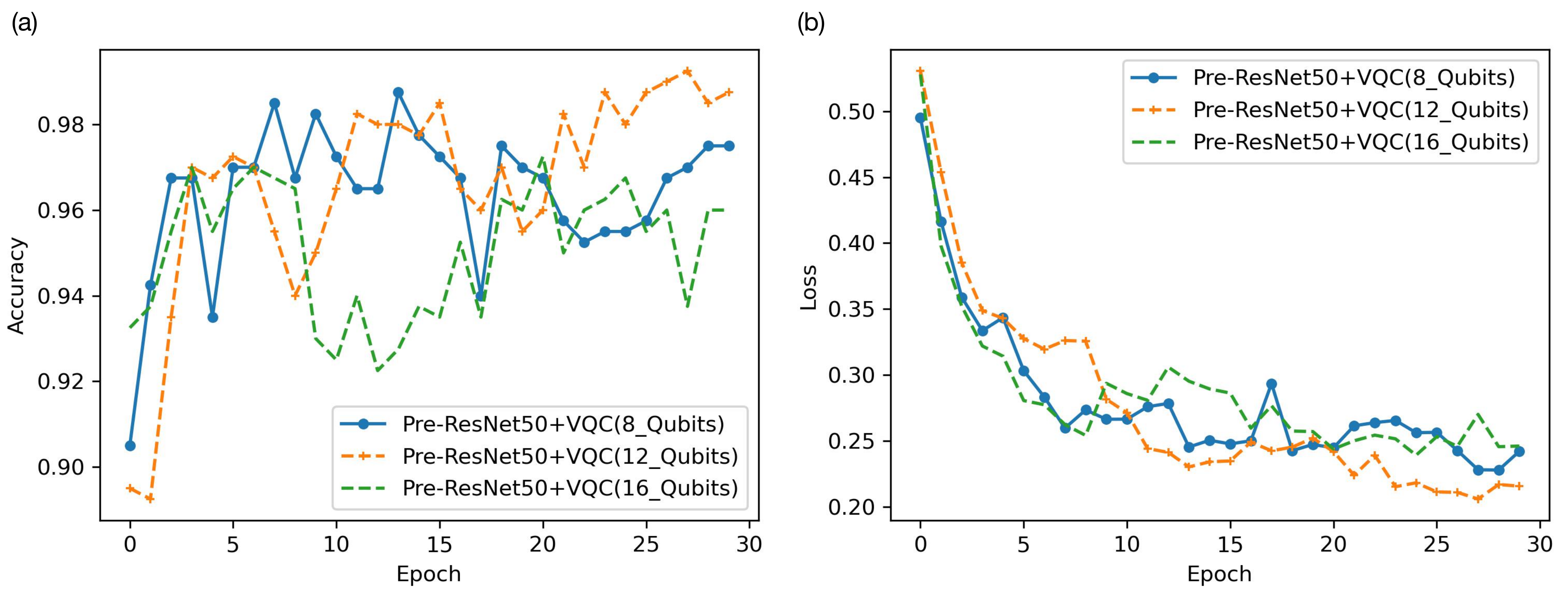, width=140mm}}
\caption{\textbf{Experimental results on clean charge stability diagrams for assessing VQC models with different numbers of qubits}. (a) The experimental results are based on accuracy; (b) The measurement of the models' performance is conducted through the loss values. We found that increasing qubits cannot consistently lead to the better empirical performance of Pre-ResNet50+VQC.}
\label{fig:dot_qubits}
\end{figure}

Additionally, we increase the number of qubits to conduct the experiments and assess the models' representation power on the clean data. As shown in Figure~\ref{fig:dot_qubits}, we witness that Pre-ResNet$50$+VQC with $12$ qubits can outperform the model with $8$ qubits, but the model with $16$ qubits cannot achieve better results than the other counterparts. The experimental results corroborate our theoretical results that the pre-trained neural networks can alleviate the dependence of qubits for the representation power. 

\begin{figure}
\centerline{\epsfig{figure=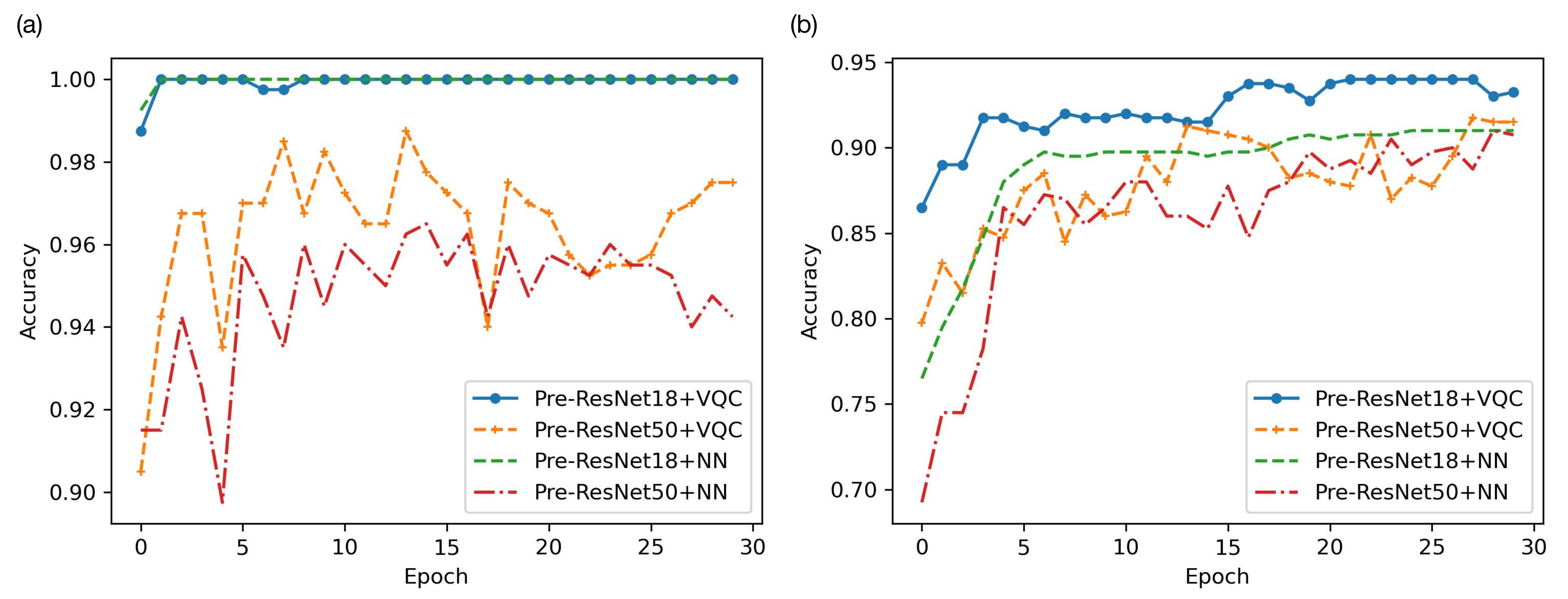, width=140mm}}
\caption{\textbf{Experimental results of pre-trained ResNet models with a VQC with $8$ qubits and a classical neural network counterpart regarding accuracy}. (a) The models' empirical results are associated with the representation power; (b) The models' empirical results correspond to the generalization power. Our empirical simulations show that the Pre-ResNet18+VQC and Pre-ResNet50+VQC can outperform their classical counterparts, Pre-ResNet18+NN and Pre-ResNet50+NN counterparts.}
\label{fig:dot_regular}
\end{figure}

Finally, as shown in Figure~\ref{fig:dot_regular}, we compare the empirical performance of Pre-ResNet18+VQC and Pre-ResNet50+VQC with their classical counterparts, Pre-ResNet18+NN and Pre-ResNet50+NN, respectively. We set the dimension of the hidden layer of NN as $8$, matching with the number of qubits $8$ used for VQC. Our empirical results show that both Pre-ResNet$18$+VQC and Pre-ResNet$50$+VQC can separately outperform their classical counterparts, Pre-ResNet18+NN and Pre-ResNet50+NN, which demonstrates that our proposed approach can be more suitable for quantum data processing. 

\subsection{Empirical results of genome TFBS prediction}

We set up a new task of genome TFBS prediction to verify our QML approach and highlight its empirical advantages. The TFBS prediction aims to build a predictive model from the data that predicts whether a DNA sequence has a specific TFBS pattern. This task can be cast as a binary classification of quantum machine learning. In this task, we use TF JunD (https://www.ebi.ac.uk/interpro/entry/ InterPro/IPR029823), a TF activator protein-1 family member that plays a central role in regulating gene transcriptions in various biological processes. 

\begin{figure}
\centerline{\epsfig{figure=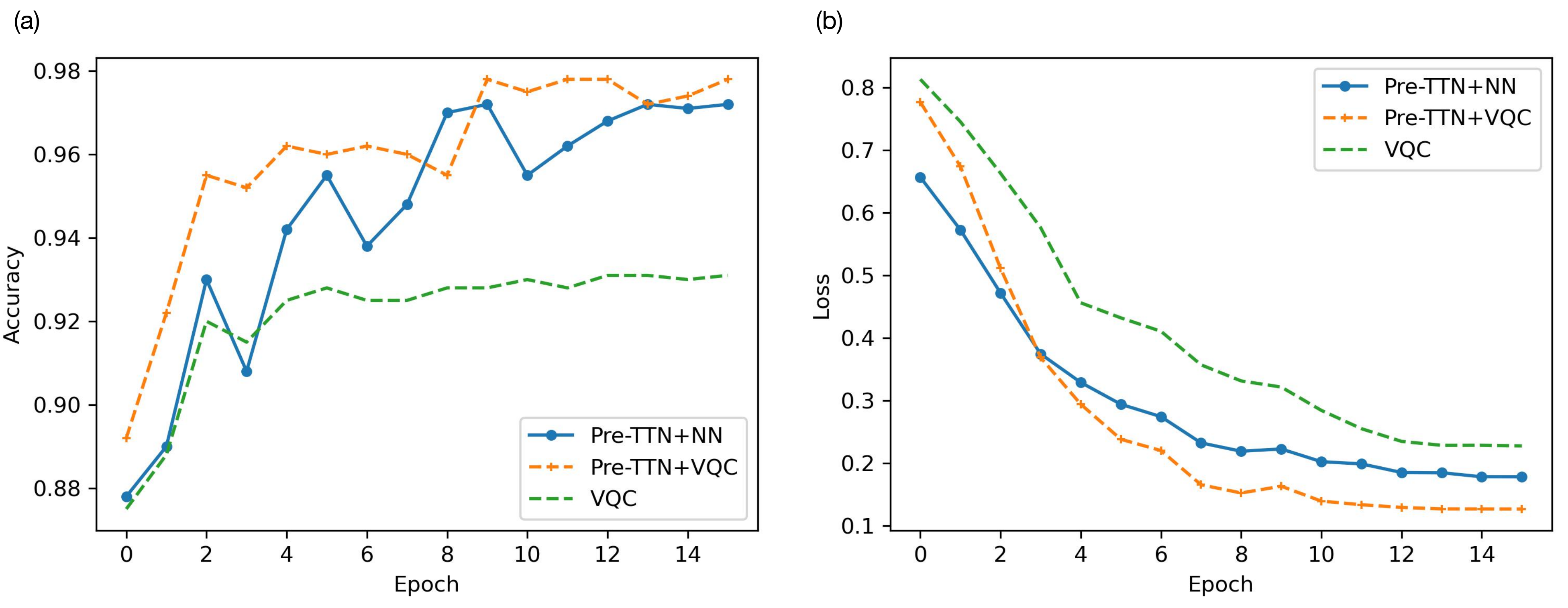, width=140mm}}
\caption{\textbf{Experimental results of Pre-TTN+VQC, VQC, and Pre-TTN+NN regarding accuracy and loss values}. (a) Comparison of the models' generalization performance in accuracy; (b) Comparison of the models' generalization performance regarding loss functions. Our empirical simulations show that the Pre-TTN+VQC can outperform its classical counterparts while enhancing the generalization performance of VQC.}
\label{fig:genome}
\end{figure}

The data are obtained from a national experiment to identify every place in the human DNA sequences where TF JunD binds. The entire $22$ chromosomes are split into segments of $101$ bases long, and each segment has been labeled to indicate whether or not it includes the JunD binding site, a specific TFBS pattern. Since each base has $4$ categories A, C, T, and G, each DNA segment is associated with $404$ dimensional features. 

We employ a tensor-train network (TTN)~\cite{oseledets2011tensor} as the pre-trained neural network $\mathcal{X}$ to form a hybrid quantum-classical model Pre-TTN+VQC, where the pre-trained TTN is obtained from our previous image and speech experiments~\cite{qi2023exploiting}, and VQC is equipped with $8$ qubits. We compare the models' performance with VQC and Pre-TTN+NN. The classical counterpart, Pre-TTN+NN, replaces VQC with a neural network with the same model parameters as VQC. Our experiments focus on the generalization power that jointly considers both representation and generalization powers, which are used to verify the advantages of our QML approach. 

Figure~\ref{fig:genome} shows the empirical comparison of the models' generalization performance on the TFBS prediction task. Our results suggest that the pre-trained TTN can attain substantial gains in the VQC's generalization performance while outperforming its classical counterpart, Pre-TTN+VQC, regarding higher accuracy and lower losses. The empirical results on the genome prediction task further corroborate our theoretical analysis that a pre-trained neural network can enhance the VQC's generalization power and obtain better empirical results on quantum datasets.

\section{Discussion}

This work demonstrates the theoretical and experimental benefits of pre-trained neural networks for the VQC block, showcasing improved representation and generalization powers. We first leverage the theoretical error performance analysis to derive upper bounds on the approximation error, estimation error, and optimization error for the Pre-$\mathcal{X}$+VQC. Our theoretical results suggest that the approximation error does not rely on the number of qubits. Instead, it corresponds to the model complexity of pre-trained neural networks and the number of source training data, which can be scaled to attain a lower approximation error. Besides, the estimation and optimization errors are jointly related to the generalization power, mainly determined by the VQC's model complexity, which is associated with the number of qubits and target data. In particular, our QML approach can lead to an exponential convergence rate in the VQC's training process, making the training of VQC easier without the PL condition. 

We conduct two binary classification tasks with target quantum data to corroborate our theoretical results. Regarding charge stability diagrams in semiconductor quantum dots, we highlight that pre-trained ResNet models can lead to better VQC representation and generalization powers, even achieving better empirical performance than their classical counterparts. As for the TFBS prediction experiments, we demonstrate that a pre-trained TTN model brings empirical performance gain in the generalization capability. The two experimental tasks corroborate our theoretical understanding of our QML approach. Although the performance gain becomes relatively marginal compared to its classical counterpart, the pre-trained classical neural networks can significantly improve the VQC's empirical performance. 

Prior works, such as a theoretical analysis of the pre-trained neural networks for VQC and its comprehensive experiments with quantum data, have yet to be delivered. Our theoretical understanding of the proposed QML approach lays a solid foundation for quantum machine learning use cases, particularly in the era of generative artificial intelligence, where pre-trained deep learning models, like large language models, are widely utilized as foundation models. Furthermore, we discuss the following issues in detail to highlight the significance of this work. 

\subsection{Which error component is dominant?}

Our previous theoretical work~\cite{qi2023theoretical} of the VQC's error performance analysis shows that a large target dataset leads to an arbitrarily small estimation error, making the approximation and optimization errors dominate the error components. 

In this work, as for the upper bound on the approximation error, we demonstrate that the pre-trained neural networks for VQC can replace the reliance on the number of qubits with a term related to the pre-trained neural network's complexity and the amount of source training data. By scaling the amount of source training data and the pre-trained neural network's complexity, we can reduce the approximation error to a small value. 

On the other hand, regarding the optimization error, we alleviate the PL condition for the VQC setup by providing an upper bound on the assumptions of approximation linearity and gradient bound. We also achieve exponential convergence rates in the training stage by controlling the constant constraints and delicately setting the learning rates. 

\subsection{The significance of our theoretical and empirical contributions}

This work focuses on QML, an interplay between machine learning and quantum computing. In the era of generative artificial intelligence, we demonstrate that using pre-trained neural networks can benefit the VQC's representation and generalization powers, which guarantees practical use cases of hybrid quantum-classical models. Dealing with quantum data like semiconductor quantum dots and TFBS prediction in human genomes is beneficial. Our contributions in this work lay a solid theoretical foundation for QML in the NISQ era while providing new experimental attempts on quantum data. They open up a new research path for QML for quantum technologies in the context of pre-trained generative artificial intelligence models for scientific computations, such as materials, medicine, mimetics, and other multidisciplinary motivations. 

\section{Methods}

This section supports the theoretical results in the main text, including the pre-trained ResNet models, TTN, cross-entropy loss, and PCA. In the Appendix, we provide a sketch of proofs for theorems. 

\subsection{Pre-trained ResNet models}

ResNet stands for Residual Neural Network, a robust convolutional neural network architecture~\cite{lawrence1997face} employed in deep learning, particularly for computer vision tasks. In particular, ResNet18 and ResNet50 are two pre-trained ResNet architectures that differ in size and complexity. ResNet18 is a smaller and faster model with around $11$ million parameters, which is a good choice when computational resources are limited or when an efficient model for inference is needed. ResNet50 is a more complex model with around $25$ million parameters, which can be generalized to diverse data scenarios but requires more computational power and training time. 

\subsection{Tensor-train network}

Tensor-train network (TTN)~\cite{oseledets2011tensor}, also known as matrix product state~\cite{cirac2021matrix}, is a type of tensor network architecture used in various scientific fields, particularly machine learning and quantum physics. They excel at representing high-dimensional tensors in a compressed and efficient manner. The TTN factorizes a high-dimensional tensor into a product of lower-dimensional tensors, typically matrices or three-index tensors, arranged in a specific chain-like structure. 

\subsection{Cross-entropy loss function}

The cross-entropy loss function~\cite{zhang2018generalized} is a popular tool used in machine learning for classification problems. It helps us measure a classification model's performance by indicating the difference between the predicted probabilities and the actual labels. The cross-entropy loss is often used with gradient descent optimization algorithms to train machine learning models. The model learns to adjust its internal parameters to better predict class labels by minimizing the loss function. In this work, the cross-entropy loss is taken as the loss function to evaluate the VQC and Pre-$\mathcal{X}$ +VQC's performance and provide gradients to update the VQC's parameters.

\subsection{Principal component analysis}

Principal component analysis (PCA)~\cite{ringner2008principal} is used in machine learning and data analysis for dimensionality reduction. It tackles datasets with many features by transforming them into smaller ones that capture the most vital information. In particular, principal components are created to capture the most significant variations in the original data. The first principal component captures the direction of the most considerable variance; the second one captures the remaining variance orthogonal to the first one, and so on. In this work, the technique of PCA is used to transform high-dimensional classical data into low-dimensional ones by adapting to the number of quantum channels used in VQC.

\section{DATA AVAILABILITY}
The dataset used in our experiments of quantum dot classification can be downloaded via the website: https://gitlab.com/QMAI/mlqe$\_$2023$\_$edx, and the dataset for TFBS predictions can be accessed via https://www.ebi.ac.uk/interpro/entry/InterPro/IPR029823.

\section{CODE AVAILABILITY}
Our codes of VQC and pre-trained classical models for VQC can be accessed through the website: https://github.com/jqi41/QuantumDot.  

\section{References}

\bibliographystyle{IEEEbib}
\bibliography{sn-bibliography}

\section{COMPETING INTERESTS}
The authors declare no Competing Financial or Non-Financial Interests.

\section{ADDITIONAL INFORMATION}
The views expressed in this article are those of the authors and do not represent the views of Wells Fargo. This article is for informational purposes only. Nothing contained in this article should be construed as investment advice. Wells Fargo makes no express or implied warranties and disclaims all legal, tax, and accounting implications related to this article.
\textbf{Correspondence} and requests for materials should be addressed to Dr. Jun Qi and Prof. Jesper Tegner.

\section*{Appendix}

\subsection*{Sketch of the proof of the main theorems}

\textbf{Proof of Theorem 1}. Given a target function $h_{\mathcal{D}}^{*}$, we define a pre-trained classical neural network $\hat{f}_{\mathcal{X}}$ and an optimal VQC operator $f^{*}_{V}$, the expected loss $\mathcal{L}_{\mathcal{D}}(f^{*}_{\mathcal{X}+V}) =  f^{*}_{V} \circ \hat{f}_{\mathcal{X}}$ is denoted as: 
\begin{equation}
\label{eq:pr11}
\epsilon_{\rm app} = \mathcal{L}_{\mathcal{D}}(f^{*}_{\mathcal{X}+V}) = \textbf{E}_{\textbf{x} \sim \mathbb{P}}\left[ \ell\left(	f_{\mathcal{V}}^{*} \circ \hat{f}_{\mathcal{X}}(\textbf{x}), h_{\mathcal{D}}^{*}(\textbf{x})	\right) \right]. 
\end{equation}

The above equation can be further decomposed into the sum of two terms: 

\begin{equation}
\begin{split}
 \epsilon_{\rm app} &= \underbrace{\textbf{E}_{\textbf{x}\sim \mathbb{P}}\left[ \ell\left( f_{V}^{*} \circ \hat{f}_{\mathcal{X}}(\textbf{x}),  f_{V}^{*} \circ f_{\mathcal{X}}^{*}(\textbf{x}) \right)\right]}_{\textcolor{blue}{\text{Term 1}}} +  \\
& \hspace{12mm} \underbrace{ \left(\textbf{E}_{\textbf{x}\sim \mathbb{P}}\left[ \ell\left(f_{V}^{*}\circ \hat{f}_{\mathcal{X}}(\textbf{x}), h_{\mathcal{D}}^{*}(\textbf{x}) \right)\right]  - \textbf{E}_{\textbf{x}\sim \mathbb{P}}\left[ \ell\left(f_{V}^{*} \circ \hat{f}_{\mathcal{X}}(\textbf{x}), f_{V}^{*} \circ f_\mathcal{X}^{*}(\textbf{x}) \right)\right]\right)}_{\textcolor{blue}{\text{Term 2}}}. 
\end{split}
\end{equation}
where we take the cross-entropy as the loss function $\ell$ in Term $1$ and Term $2$, then we can further derive that 

\begin{equation}
\begin{split}
\text{Term $1$} &= \textbf{E}_{\textbf{x}\sim \mathbb{P}}\left[ \ell\left(f_{V}^{*} \circ  \hat{f}_{\mathcal{X}}(\textbf{x}),  f_{V}^{*} \circ f_{\mathcal{X}}^{*}(\textbf{x}) \right)\right]  \\ 
&= \textbf{E}_{\textbf{x} \sim \mathbb{P}} \left[ \left(- f_{V}^{*} \circ \hat{f}_{\mathcal{X}}(\textbf{x}) \right) \log \left(\frac{f^{*}_{V} \circ f_{\mathcal{X}}^{*}(\textbf{x})}{f_{V}^{*} \circ \hat{f}_{\mathcal{X}}(\textbf{x})} \right) \right]  \\
&\le \textbf{E}_{\textbf{x} \sim \mathbb{P}} \left\vert \left(-f_{V}^{*} \circ \hat{f}_{\mathcal{X}}(\textbf{x}) \right) \left(	\frac{f^{*}_{V} \circ f_{\mathcal{X}}^{*}(\textbf{x})}{f_{V}^{*} \circ \hat{f}_{\mathcal{X}}(\textbf{x})}  -  1	\right)  \right\vert \\
&= \textbf{E}_{\textbf{x} \sim \mathbb{P}} \vert f_{V}^{*} \circ \hat{f}_{\mathcal{X}}(\textbf{x}) - f_{V}^{*} \circ \hat{f}_{\mathcal{X}}(\textbf{x}) \vert \\
&\le \textbf{E}_{\textbf{x} \sim \mathbb{P}} \sup\limits_{f_{\mathcal{X}} \in \mathbb{F}_{\mathcal{X}}} \vert f_{V}^{*} \circ  \hat{f}_{\mathcal{X}}(\textbf{x}) - f_{V}^{*}\circ f_{\mathcal{X}}(\textbf{x}) \vert
\end{split}
\end{equation}

The last term can be further upper bounded with an empirical Rademacher complexity $\hat{\mathcal{R}}_{S}(\mathcal{X})$, which is related to the upper bound as $\tilde{\mathcal{O}} \left( \sqrt{\frac{C(\mathbb{F}_{\mathcal{X}})}{\vert D_{A} \vert}} \right)$, where $C(\mathbb{F}_{\mathcal{X}})$ represents a measurement of the intrinsic complexity of the functional class $\mathbb{F}_{\mathcal{X}}$, and $D_{A}$ refers to the source dataset. Thus, we can attain the upper bound as: 

\begin{equation}
\text{Term $1$} = \textbf{E}_{\textbf{x}\sim \mathbb{P}}\left[ \ell\left(f_{V}^{*} \circ  \hat{f}_{\mathcal{X}}(\textbf{x}),  f_{V}^{*} \circ f_{\mathcal{X}}^{*}(\textbf{x}) \right)\right] = \tilde{\mathcal{O}} \left( \sqrt{\frac{C(\mathbb{F}_{\mathcal{X}})}{\vert D_{A} \vert}} \right). 
\end{equation}

On the other hand, 

\begin{equation}
\begin{split}
\text{Term $2$} &= \textbf{E}_{\textbf{x}\sim \mathbb{P}} \left[\ell\left(f_{V}^{*} \circ  \hat{f}_{\mathcal{X}}(\textbf{x}), h_{\mathcal{D}}^{*}(\textbf{x}) \right)\right]  - \textbf{E}_{\textbf{x}\sim \mathbb{P}}\left[ \ell\left(f_{V}^{*} \circ \hat{f}_{\mathcal{X}}(\textbf{x}), f_{V}^{*} \circ f_{\mathcal{X}}^{*}(\textbf{x}) \right)\right] \\
&= \textbf{E}_{\textbf{x} \sim \mathbb{P}} \left[-f_{V}^{*} \circ \hat{f}_{\mathcal{X}}(\textbf{x}) \log\left( \frac{h_{\mathcal{D}}^{*}(\textbf{x})}{f_{V}^{*} \circ \hat{f}_{\mathcal{X}}(\textbf{x})} \right) +  f_{V}^{*}\circ \hat{f}_\mathcal{X}(\textbf{x}) \log \left(\frac{f_{V}^{*} \circ f_{\mathcal{X}}^{*}(\textbf{x})}{f_{V}^{*} \circ \hat{f}_{\mathcal{X}}(\textbf{x})} \right)  \right] \\
&\le \textbf{E}_{\textbf{x} \sim \mathbb{P}} \left\vert  f_{V}^{*} \circ \hat{f}_{\mathcal{X}}(\textbf{x}) \left( \log\left(f_{V}^{*} \circ f_{\mathcal{X}}^{*}(\textbf{x})) - \log(h_{\mathcal{D}}^{*}(\textbf{x})\right)\right) \right\vert. 
\end{split}
\end{equation}

Since the quantum circuit $f_{V}^{*}$ can be expressed as an exponential form, based on the universal approximation theory of quantum neural networks, we can further derive that 
\begin{equation}
\vert \log(f_{V}^{*} \circ f_{\mathcal{X}}^{*}(\textbf{x})) - \log(h_{\mathcal{D}}(\textbf{x})) \vert \le \frac{1}{\sqrt{M}}. 
\end{equation}

Then, we finally derive that 
\begin{equation}
\text{Term $2$} \le \textbf{E}_{\textbf{x} \sim \mathbb{P}} \left\vert f_{V}^{*} \circ \hat{f}_{\mathcal{X}}(\textbf{x}) \left( \log\left(f_{V}^{*} \circ f_{\mathcal{X}}^{*}(\textbf{x})) - \log(h_{\mathcal{D}}^{*}(\textbf{x})\right)\right) \right\vert = \mathcal{O}\left( \frac{1}{\sqrt{M}} \right). 
\end{equation}

\noindent \textbf{Proof of Theorem 2}. Given the target dataset $D_{B}$ for the VQC's training, the estimation error $\epsilon_{\rm est}^{'}$ is only related to the VQC's model complexity because we aim to update the VQC's model parameters and freeze the pre-trained model's parameters. Then, we can directly derive that 
\begin{equation}
\epsilon_{\rm est} = \tilde{\mathcal{O}}\left( \sqrt{\frac{C(\mathbb{F}_{V})}{\vert D_{B} \vert}} \right). 
\end{equation}

\noindent \textbf{Proof of Theorem 3}. 
Please assume that the VQC's parameters are represented as $\boldsymbol{\theta}$; we denote $\boldsymbol{\theta}_{t}$ as the VQC's parameters at the epoch $t$ during its training stage, which iteratively approximates the optimal parameters $\boldsymbol{\theta}^{*}$. Given the target dataset $D_{B}$ and a learning rate $\eta$, we use the empirical loss $\mathcal{L}_{S}(\cdot)$ to approximate the expected loss $\mathcal{L}_{\mathcal{D}}(\cdot)$ such that an SGD algorithm returns 

\begin{equation}
\begin{split}
r_{t+1}^{2} &= \lVert  \boldsymbol{\theta}_{t+1} - \boldsymbol{\theta}^{*} \rVert_{2}^{2} \\ 
&\le \lVert 	\boldsymbol{\theta}_{t} - \eta \nabla\mathcal{L}_{S}(\boldsymbol{\theta}_{t}) - \boldsymbol{\theta}^{*}	\rVert_{2}^{2} \\
&= r_{t}^{2} + 2\eta \langle \nabla_{\boldsymbol{\theta}}\mathcal{L}_{S}(\boldsymbol{\theta}_{t}), \boldsymbol{\theta}^{*} - \boldsymbol{\theta}_{t} \rangle + \eta^{2} \lVert \nabla_{\boldsymbol{\theta}}\mathcal{L}_{S}(\boldsymbol{\theta}_{t}) \rVert_{2}^{2}. 
\end{split}
\end{equation}

To further upper bound $r_{t+1}^{2}$, given the constants $\beta, R, L$ defined in Assumptions 1 and 2, we employ Assumptions 1 and 2 to separately upper bound $\langle \nabla_{\boldsymbol{\theta}}\mathcal{L}_{S}(\boldsymbol{\theta}_{t}), \boldsymbol{\theta}^{*} - \boldsymbol{\theta}_{t} \rangle$ and $\lVert \nabla_{\boldsymbol{\theta}}\mathcal{L}(\boldsymbol{\theta}_{t}) \rVert_{2}^{2}$ as: 

\begin{enumerate}
\item To upper bound the term $\langle \nabla_{\boldsymbol{\theta}}\mathcal{L}_{S}(\boldsymbol{\theta}_{t}), \boldsymbol{\theta}^{*} - \boldsymbol{\theta}_{t} \rangle$, based on Assumption 1 regarding the approximate convexity theory for the VQC's parameters, we have 
\begin{equation}
\langle \nabla_{\boldsymbol{\theta}}\mathcal{L}_{S}(\boldsymbol{\theta}_{t}), \boldsymbol{\theta}^{*} - \boldsymbol{\theta}_{t} \rangle \le \mathcal{L}_{S}(\boldsymbol{\theta}^{*}) - \mathcal{L}_{S}(\boldsymbol{\theta}_{t}) + 4\beta R^{2}.  
\end{equation}

\item To upper bound the term $\lVert \nabla_{\boldsymbol{\theta}}\mathcal{L}_{S}(\boldsymbol{\theta}_{t}) \rVert_{2}^{2}$, based on Assumption 2 about the gradient bound on the VQC's parameters, we have 

\begin{equation}
\lVert \nabla_{\boldsymbol{\theta}}\mathcal{L}_{S}(\boldsymbol{\theta}_{t}) \rVert_{2}^{2} \le L^{2} + \beta^{2} R^{2}. 
\end{equation}
\end{enumerate}

Therefore, at the epoch $t+1$ of total iterations $T_{\rm sgd}$, we derive that 

\begin{equation}
\begin{split}
r_{t+1}^{2} &\le r_{t}^{2} + 2\eta \left( \mathcal{L}(\boldsymbol{\theta}^{*}) - \mathcal{L}_{S}(\boldsymbol{\theta}_{t}) + 4\beta R^{2} \right) + \eta^{2} (L^{2} + \beta^{2} R^{2}) \\
&\le r_{0} + 2\eta \left( \sum\limits_{t=0}^{T_{\rm sgd}-1} \mathcal{L}_{S}(\boldsymbol{\theta}^{*}) - \mathcal{L}_{S}(\boldsymbol{\theta}_{t}) \right) + 8\eta T_{\rm sgd} \beta R^{2} + \eta^{2} T_{\rm sgd} (L^{2} + \beta^{2} R^{2}). 
\end{split}
\end{equation}

By rearranging the last inequality, we have 

\begin{equation}
\begin{split}
\frac{r_{0}^{2} - r_{T_{\rm sgd}}^{2}}{2\eta T_{\rm sgd}} + 4\beta R^{2} + \frac{\eta}{2} \left( L^{2} + \beta^{2} R^{2} \right) &\ge \frac{1}{T_{\rm sgd}} \sum\limits_{t=0}^{T_{\rm sgd}-1} \mathcal{L}_{S}(\boldsymbol{\theta}^{*}) - \mathcal{L}_{S}(\boldsymbol{\theta}_{t}) \\ 
&\ge \min\limits_{t=0,...,T_{\rm sgd}-1} \mathcal{L}_{S}(\boldsymbol{\theta}^{*}) - \mathcal{L}_{S}(\boldsymbol{\theta}_{t}). 
\end{split}
\end{equation}

Now, using the step size $\eta = \frac{1}{\sqrt{T_{\rm sgd}}}\left( \frac{R}{\sqrt{L^{2} + \beta^{2} R^{2}}} \right)$, we observe that 

\begin{equation}
\begin{split}
\frac{r_{0}^{2} - r_{T_{\rm sgd}}^{2}}{2\eta T_{\rm sgd}} + \frac{\eta}{2}\left( L^{2} + \beta^{2} R^{2} \right) \le \frac{R^{2}}{2\eta T_{\rm sgd}} + \frac{\eta}{2} \left(L^{2} + \beta^{2} R^{2} \right)  = R \sqrt{\frac{L^{2} + \beta^{2} R^{2}}{T_{\rm sgd}}}, 
\end{split}
\end{equation}
from which the upper bound for the optimization error in Theorem 3 follows, where 

\begin{equation}
\epsilon_{\rm opt} \le \beta R^{2} + R \sqrt{\frac{L^{2} + \beta^{2} R^{2}}{T_{\rm sgd}}}. 
\end{equation}

\end{document}